\documentclass[letterpaper]{article} 
\usepackage{aaai2026}  
\usepackage{times}  
\usepackage{helvet}  
\usepackage{courier}  
\usepackage[hyphens]{url}  
\usepackage{graphicx} 
\usepackage{natbib}  
\usepackage{caption} 
\usepackage{booktabs}
\usepackage{algorithm}
\usepackage{algorithmic}
\usepackage{amsfonts}
\usepackage{amsmath}
\usepackage{multirow}
\usepackage{newfloat}
\usepackage{listings}
\DeclareCaptionStyle{ruled}{labelfont=normalfont,labelsep=colon,strut=off} 
\floatstyle{ruled}
\newfloat{listing}{tb}{lst}{}
\floatname{listing}{Listing}
\title{Minimal High-Resolution Patches Are Sufficient for Whole Slide Image Representation via Cascaded Dual-Scale Reconstruction}

\author{
    Yujian Liu\textsuperscript{\rm 1}\equalcontrib,
    Yuechuan Lin\textsuperscript{\rm 1}\equalcontrib,
    Dongxu Shen\textsuperscript{\rm 2}\equalcontrib,
    Haoran Li\textsuperscript{\rm 1},
    Yutong Wang\textsuperscript{\rm 1},\\
    Qingquan Wang\textsuperscript{\rm 1},
    Xiaoli Liu\textsuperscript{\rm 3}\thanks{Corresponding author},
    Shidang Xu\textsuperscript{\rm 1}\footnotemark[2]
}
\affiliations{
    South China University of Techenology\textsuperscript{\rm 1}\\

    Hong Kong University of Science and Technology (Guangzhou)\textsuperscript{\rm 2}\\

    AiShiWeiLai AI Research, Beijing, China\textsuperscript{\rm 3}\\

}

\begin{document}

\maketitle

\begin{abstract}
Whole-slide image (WSI) analysis remains challenging due to the gigapixel scale and sparsely distributed diagnostic regions. Multiple Instance Learning (MIL) mitigates this by modeling the WSI as bags of patches for slide-level prediction. However, most MIL approaches emphasize aggregator design while overlooking the impact of the feature extractor of the feature extraction stage, which is often pretrained on natural images. This leads to domain gap and suboptimal representations. Self-supervised learning (SSL) has shown promise in bridging domain gap via pretext tasks, but it still primarily builds upon generic backbones (e.g., ResNet, ViT), thus requiring WSIs to be split into small patches. This inevitably splits histological structures and generates both redundant and interdependent patches, which in turn degrades aggregator performance and drastically increases training costs. To address this challenge, we propose a Cascaded Dual-Scale Reconstruction (CDSR) framework, demonstrating that only an average of 9 high-resolution patches per WSI are sufficient for robust slide-level representation. CDSR employs a two-stage selective sampling strategy that identifies the most informative representative regions from both model-based and semantic perspectives. These patches are then fed into a Local-to-Global Network, which reconstructs spatially coherent high-resolution WSI representations by integrating fine-grained local detail with global contextual information. Unlike existing dense-sampling or SSL pipelines, CDSR is optimized for efficiency and morphological fidelity. Experiments on Camelyon-16, TCGA-NSCLC, and TCGA-RCC demonstrate that CDSR achieves improvements of 6.3\% in accuracy and 5.5\% in area under ROC curve on downstream classification tasks with only 7,070 high-resolution patches per dataset on average, outperforming state-of-the-art methods trained on over 10,000,000 patches. 

\begin{figure}[t]
\centering
\includegraphics[width=1\columnwidth]{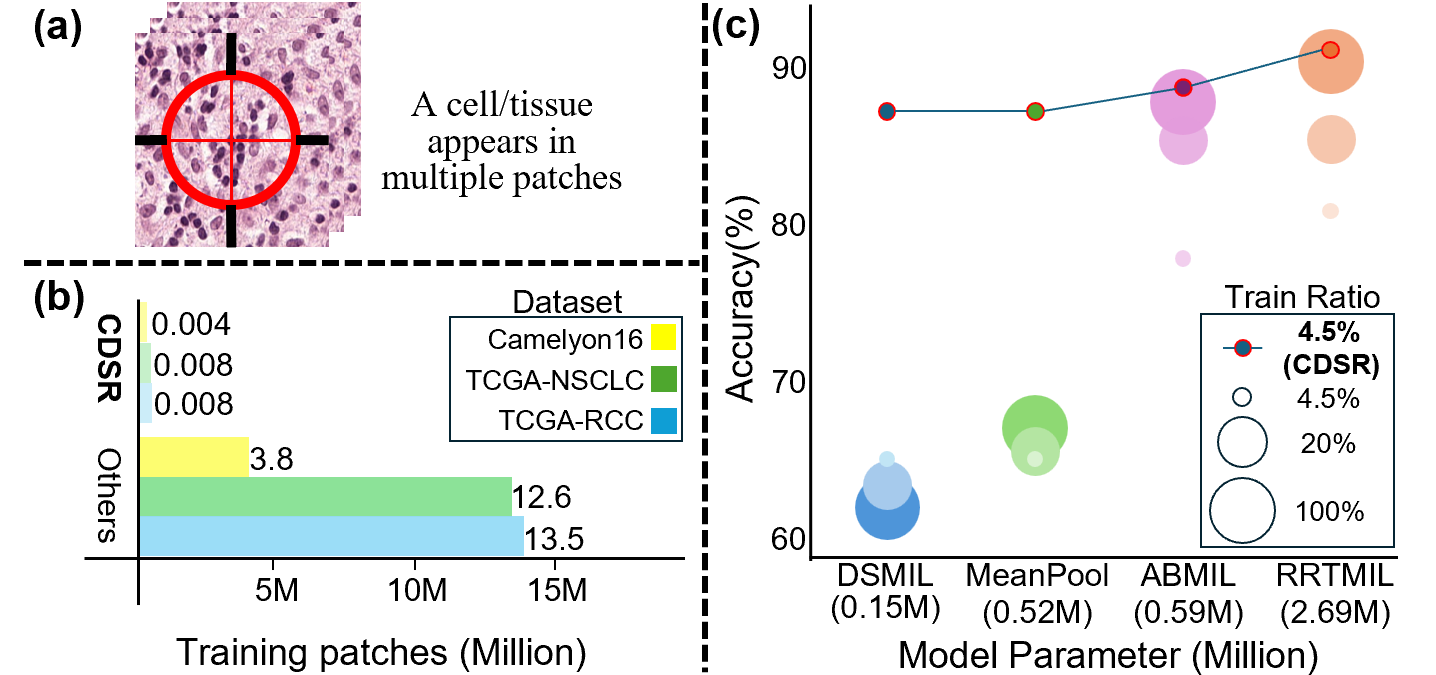} 
\caption{Motivation and superiority of the proposed Cascaded Dual-Scale Reconstruction (CDSR) framework. \textbf{(a)} Small patch division in MIL often disrupts critical structures like cells.
\textbf{(b)} CDSR significantly reduces training patches and computational cost compared to self-supervised learning.
\textbf{(c)} With only 4.5\% of the total training data, CDSR outperforms existing methods in classification accuracy.}
    \label{fig:figure1}
\end{figure}

\end{abstract}
\section{Introduction}
Whole slide images (WSIs) are gigapixel-scale scans of histopathological tissue and have become fundamental for computer-aided diagnosis in pathology \cite{wang2022scl,qu2022bi,gurcan2009histopathological}. However, their immense size poses significant difficulties, as they typically include extensive homogeneous regions, large uninformative areas, and small but diagnostically important regions \cite{srinidhi2021deep,LongMIL,WiKG}.

To address these issues, Multiple Instance Learning (MIL) has emerged as a standard paradigm, treating each WSI as a bag of image patches \cite{mercan2016multi,campanella2019clinical}. However, this pipeline often results in suboptimal representations due to two key limitations. First, WSIs are typically partitioned into thousands of small, fragmented important histological structures \cite{hou2016patch} such as glands, tumor boundaries, and stromal patterns, disrupting spatial continuity \cite{HIPT,raghu2019transfusion} (Fig.~\ref{fig:figure1}(a)). Second, most MIL frameworks adopt feature extractors pretrained on natural image datasets \cite{deng2009imagenet,canziani2016analysis}, which introduces a pronounced domain gap between natural training features and medical imaging characteristics \cite{konz2024effect,morra2021bridging}. Consequently, the aggregator network is forced to operate on noisy, redundant, and often weakly informative embeddings, compromising overall performance.

Self-supervised learning (SSL) has emerged as a promising strategy to reduce the domain gap by learning visual representations from unlabeled medical images via pretext tasks such as rotation prediction or image reconstruction \cite{bucci2021self,chen2020simple,he2020momentum,rani2023self}. However, existing SSL frameworks still rely heavily on standard vision backbones \cite{Caron} (e.g., ResNet \cite{he2016deep}, ViT \cite{dosovitskiy2020image}), which are originally designed for small-sized natural images. Therefore, WSI must be divided into millions of small patches, leading to three major issues: (1) increased computational burden  \cite{grill2020bootstrap,gui2024survey} (Fig.~\ref{fig:figure1}(b)), (2) redundancy and noise due to irrelevant patches \cite{koohbanani2021self,he2022masked}, and (3) disruption of morphological continuity \cite{wang2021transpath,lazard2023giga} (Fig.~\ref{fig:figure1}(a)). These factors jointly hinder downstream tasks by feeding the model with fragmented and weakly correlated patches. Basically, the reliance on uniform, dense sampling and generic feature extractors limits the representation capacity of existing WSI pipelines. This raises a key question: Can we efficiently represent a WSI using only a small number of high-resolution patches that preserve both semantic relevance and structural integrity?

Motivated by this, we propose our Cascaded Dual-Scale Reconstruction (CDSR)—a novel framework that bypasses the need for dense patch extraction and domain-mismatched pretraining. CDSR comprises two key components: a two-stage selective sampling strategy and a Local-to-Global Network (L2G-Net). The sampling strategy identifies an average of 9 diagnostically informative patches per WSI from both model-based and semantic perspectives, dramatically reducing the number of training patches to 7,070 per dataset (4.5\% of the total training data) on average (Fig.~\ref{fig:figure1}(b, c)). This dramatically reduces computational overhead while minimizing redundancy. Critically, our sampling process ensures that selected patches capture the most morphologically and diagnostically relevant regions of interest. Next, the selected representative patches are center-expanded to higher resolution and passed into the L2G-Net to extract detailed features. By processing enlarged patches with transformer-based global context and multi-view low-resolution guidance, the L2G-Net preserves structural continuity and alleviates the fragmentation caused by dense patching strategies. This results in a unified, information-rich representation of the tissue landscape, optimized for downstream prediction tasks. In summary, our main contributions are as follows:

\begin{itemize}
    \item We introduce a two-stage selective sampling strategy that selects representative positive and negative patches from both model-based and semantic perspectives. This ensures that training patches are concentrated in diagnostically relevant regions, enabling the L2G-Net to operate with only an average of 9 patches per WSI.

    \item We propose a L2G-Net that leverages low-resolution multi-view guidance and the global receptive field of transformers to reconstruct large-size patches. This design overcomes the limitations of patch-wise feature extraction and achieves high-quality contextual reconstruction, leading to significant gains in representation quality and accuracy.
    
    \item Our proposed CDSR framework is the first to introduce a dedicated feature extractor for high-resolution patches in WSI recognition, providing a lightweight yet effective alternative to conventional dense sampling pipelines. Despite using only 4.5\% of the total training data and $17\%$ of the training time on average, it achieves improvements of 6.3\% in accuracy and 5.5\% in area under ROC curve compared to traditional MIL frameworks trained on the entire dataset (Fig.~\ref{fig:figure1}(c)).

\end{itemize}

\section{Related Work}

\subsection{Patch Selection Strategies}
\textbf{Random sampling.} 
Early MIL methods draw patches uniformly at random \cite{zhu2017wsisa}, which was subsequently extended to multi‐scale random sampling \cite{tang2025revisiting}, reducing memory load and implicitly regularizing the model. However, random schemes are lightweight but suffer from slower convergence, under‐utilized label signals, and excessive sensitivity to the randomness in rare but critical regions.

\noindent\textbf{Selective sampling.} More targeted strategies use clustering or attention scores to guide patch choice. Cluster‐to‐Conquer \cite{C2C} partitions patches into clusters and samples from each to ensure coverage of diverse discriminative features. End‐to‐End Attention Pooling methods \cite{E2E} score all patches via an attention network and select the highest and lowest scoring subsets. These selective sampling methods produce homogeneous selections and still exhibit redundancy.

\subsection{Multiple Instance Learning}

Conventional multiple instance learning (MIL) approaches for whole-slide image analysis typically employ ResNet-50 pretrained on ImageNet \cite{deng2009imagenet}, coupled with aggregation schemes. Attention‑based MIL methods \cite{AB-MIL,CLAM} proposed a trainable attention module to perform attention‑weighted pooling of patch embeddings into a compact slide‑level feature. Later approaches focus on capturing long-distance dependencies among patch embeddings, which improves the performance on correctly identifying and distinguishing relevant histopathological patterns \cite{TransMIL}.
The drawback of these methods is that they rely on frozen ResNet‑50 features and cannot adapt patch representations during training, leading to suboptimal patch features and exacerbating the domain gap.

\subsection{Self‐Supervised MIL}

To overcome the domain gap, recent works integrate self-supervised framework such as DINO \cite{dino} and SimCLR \cite{simclr} into the MIL pipeline. Dual-stream MIL (DSMIL) \cite{DSMIL} introduces a dual-stream non-local operation and employs self-supervised contrastive pretraining to refine patch embeddings. Iteratively Coupled MIL (ICMIL) \cite{ICMIL} further bridges bag and patch level learning by iteratively fine-tuning the patch encoder and incorporating a teacher–student framework. HIPT \cite{HIPT} leverages hierarchical DINO self-supervised pretraining to capture fine-grained features and global semantic context. Existing self-supervised methods focus on low-resolution patch reconstruction and require extensive processing time.

\begin{figure*}[t]
\centering
\includegraphics[width=0.9 \textwidth]{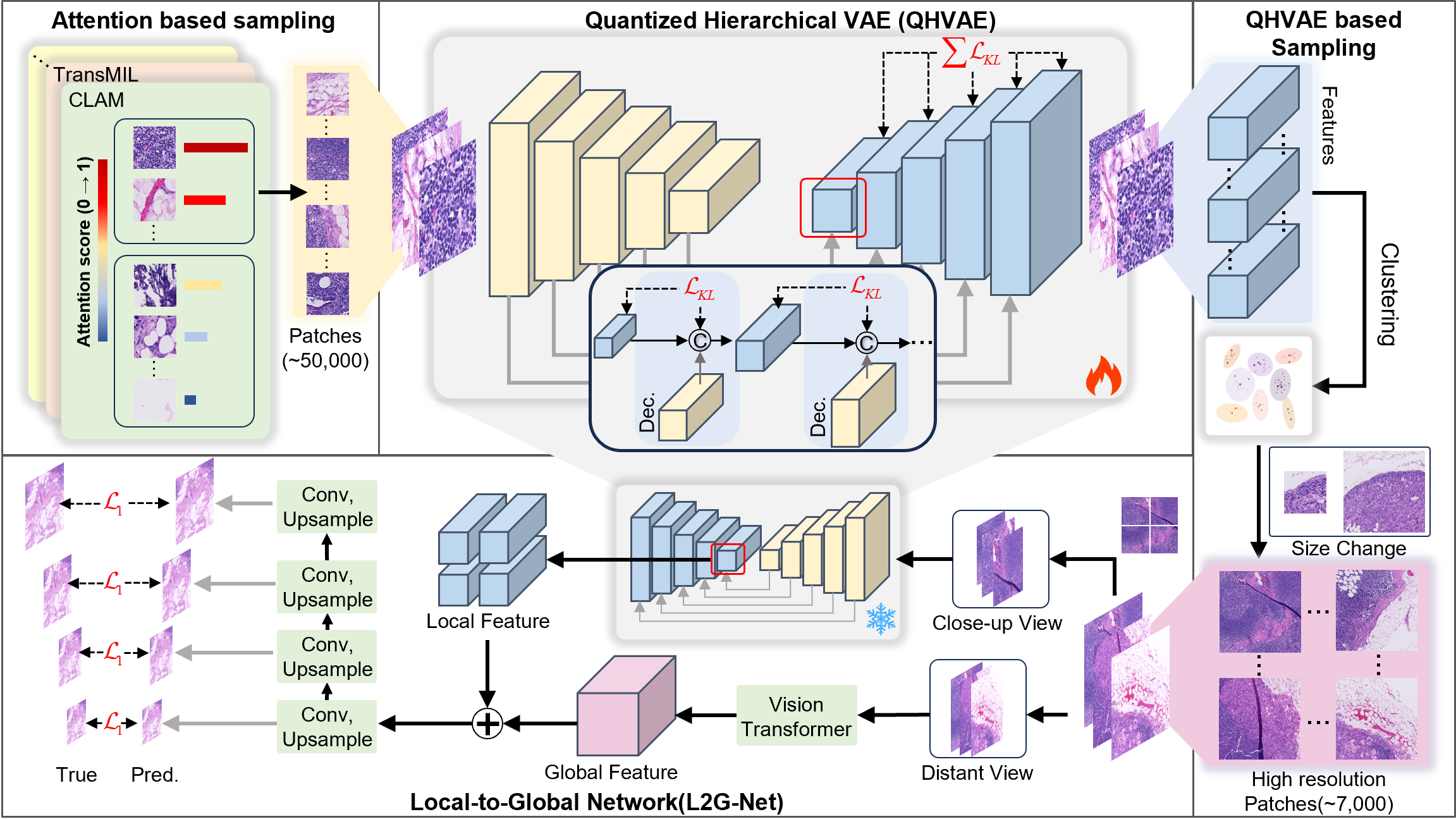} 
\caption{Overview of proposed CDSR framework. High resolution WSI patches are ranked by an ensemble of MIL attention maps, and the top‑scoring regions are forwarded to the Quantized Hierarchical VAE-based clustering model that preserves tissue diversity and removes redundancy. The resulting representative patches are then processed by Local-to-Global Network through parallel transformer-based distant-view and QHVAE-based close-up feature extraction. Finally, the fused features undergo progressive upsampling and convolutional refinement, with a hierarchical loss guiding full-resolution slide reconstruction.}
    \label{fig:pipeline}
\end{figure*}

\section{Method}

\subsection{Overview Architecture}
The Cascaded Dual-Scale Reconstruction (CDSR) framework is devised to extract latent feature representations from representative high-resolution patches. Consequently, as illustrated in Fig.~\ref{fig:pipeline}, CDSR is designed around two complementary components, a two-stage sampling that selects the most informative patches and a Local-to-Global Network (L2G-Net) that rebuilds high-resolution images.

\noindent\textbf{Two-stage Selective Sampling.}
Initially, an ensemble of advanced MIL models assigns an attention score to every patch on the slide. These scores are aggregated into a unified patch map, from which only the top-ranked patches containing histological pattern information are selected. These patches are subsequently forwarded to the Quantized Hierarchical VAE (QHVAE)-based cluster-sampling stage, which balances coverage of diverse tissue patterns. Representative patches are embedded by QHVAE, producing semantically rich feature vectors. For each slide, these embeddings are partitioned into a compact set of $k$-means clusters, from which we uniformly sample patches and feed them into the Local-to-Global Network.

\noindent\textbf{Local-to-Global Network.} After the two-stage sampler selects the most informative patches, the encoder of Local-to-Global Network extracts features from two views. A distant view leverages a Swin Transformer to capture long-range dependencies, whereas a close-up view reuses our well-trained QHVAE to extract high-quality local semantics. Subsequently, the decoder employs a series of interpolation and convolution layers to upscale the fused features of the two views. Multi-scale loss supervision guides reconstruction accuracy.

\subsection{QHVAE-Based Patch Encoding Model}
Quantized Hierarchical VAE (QHVAE) is first introduced due to its dual role as a bridge between the sampling and reconstruction stages. On the one hand, it provides informative features to guide the sampling process; on the other hand, it serves as the encoder for the close-up view in the high-resolution Local-to-Global Network (L2G-Net). To capture the semantic representations of these representative patches, QHVAE is employed to learn high-level structures in a self-supervised framework, making it especially suitable for large-scale unlabeled datasets. QHVAE is built upon the QRes-VAE backbone \cite{VAE,lossyvae}.

\noindent\textbf{Model Architecture.}
The model uses a hierarchical structure with a bottom-up encoder and a top-down decoder. The encoder downsamples \( I_2 \) into multiscale features using residual blocks. Formally, the encoder features are:
\begin{equation}
\quad h_i^{\text{enc}} = \mathrm{ConvBlock}_i(h_{i-1}^{\text{enc}}), \quad i=1,\ldots,N
\end{equation}
where \( h_i^{\text{enc}} \) denotes the encoder feature at level \( i \), and \( h_0^{\text{enc}} = I_2 \). \(\mathrm{ConvBlock}_i\) adopts the ConvNeXt block architecture with residuals.

The decoder starts from a learnable constant tensor \( r \in \mathbb{R}^{w \times 1 \times 1} \), with \( h_0^{\text{dec}} = r \). It reconstructs the image by progressively refining features through stacked latent blocks. Each block injects a latent variable and combines it with previous features via convolutions, enabling the model to capture hierarchical and compressible representations.
\begin{equation}
\begin{gathered}
z_i = \mathrm{LatentBlock}_i\!\bigl(h_{i-1}^{\text{dec}},\,h_i^{\text{enc}}\bigr), 
\quad i = 1,\ldots,N \\[4pt]
h_i^{\text{dec}} = \mathrm{ConvBlock}_i(z_i) + h_{i-1}^{\text{dec}} 
\end{gathered}
\label{eq:latent_update}
\end{equation}
where \( h_i^{\text{enc}} \) and \( h_i^{\text{dec}} \) denote the encoder and decoder features at level \( i \), respectively, \( z_i \) is the inferred latent variable, and \( h_i^{\text{dec}} \) is used in subsequent layers for reconstruction.

\noindent\textbf{Training phase.}
Each latent block is trained by minimizing the KL divergence between the approximate posterior and the prior of its corresponding latent variable. Specifically, each latent variable $z_i$ is modeled with a uniform posterior centered at a learned mean $\mu_i$:
\begin{equation}
q_i(z_i | z_{<i}, x) = \mathcal{U} \left( \mu_i - \frac{1}{2}, \mu_i + \frac{1}{2} \right)
\end{equation}
This design enables continuous latent to be effectively discretized by approximating quantization with a uniform distribution. The corresponding prior for each $z_i$ is defined as a Gaussian convolved with a unit-width uniform distribution:
\begin{equation}
p_i(z_i | z_{<i}) = \int_{z_i - \frac{1}{2}}^{z_i + \frac{1}{2}} \mathcal{N}(t; \hat{\mu}_i, \hat{\sigma}_i^2) dt
\end{equation}
Here, $\hat{\mu}_i$ and $\hat{\sigma}_i$ are predicted from previously decoded latent variables, denoted as $z_{<i}$, representing all latents preceding index $i$. The convolution with a uniform distribution ensures that the prior aligns with the quantized nature of the latent space. The QHVAE is trained by minimizing the following rate-distortion loss:
\begin{equation}
\begin{aligned}
\mathcal{L} &= D_{\mathrm{KL}}\big(q(z|x) \,\|\, p(z)\big) + \mathbb{E}_{q(z|x)} \left[ \log \frac{1}{p(x|z)} \right] \\
            &= \mathbb{E}_{q(z|x)} \left[ \sum_{i=1}^N \log \frac{1}{p_i(z_i | z_{<i})} + \lambda \cdot \|x - \hat{x}\|_2^2 \right] + c
\end{aligned}
\end{equation}
where the first term measures the information cost of encoding latent variables, and \( \|x - \hat{x}\|_2^2 \) denotes the reconstruction distortion quantified by the squared \(L_2\) norm.

\noindent\textbf{Inferencing phase.}
During compression, the latent variables are quantized via residual, where $\left\lfloor \cdot \right\rceil$ is the nearest integer rounding function:
\begin{equation}
    z_i \leftarrow \hat{\mu}_i + \left\lfloor \mu_i - \hat{\mu}_i \rceil \right.
\end{equation}
The quantized latents are then compressed into bitstreams using a Gaussian PMF with RANS-based entropy coders. During decompression, these bitstreams are decoded sequentially to recover each $z_i$, which is then incorporated into the feature hierarchy to reconstruct the image. After training, input patches are processed through the compression-decompression pipeline, and the intermediate latent features extracted during decompression are utilized for downstream tasks.

\subsection{Two-stage Selective Sampling}

Prior patch-based WSI studies analysis were hampered by redundant, non-informative patches, frequently suffered from redundancy, with a proliferation of non-informative patches. Increasing the relevance of the retained patches while suppressing irrelevant ones is therefore an effective preprocessing strategy. Our two-stage selective sampling strategy, which combines attention-score filtering with QHVAE-based clustering, enhances downstream efficiency and improves classification accuracy.

\noindent\textbf{Sampling Based on Attention Score.} Visual attention mechanisms have proven effective at pinpointing critical patterns; accordingly, we begin the pipeline with an attention-score–driven sampling stage. 
A set of representative attention-based MIL models is employed to calculate per-patch attention scores, whose target patch sizes range across large resolutions, from coarse to fine.
Specifically, let \(h_{m}^{(n)} \in \mathbb{R}^{D}\) be the feature vector of the \(m\)-th patch produced by the \(n\)-th model. The attention score \(A_{m}^{(n)}\) is calculated using a gated-tanh network followed by normalization:
\begin{equation}
A_{m}^{(n)}
=\frac{\displaystyle
\exp\!\Bigl(
w^{\mathsf T}\!\bigl[\tanh\!\bigl(V_{1}h_{m}^{(n)}\bigr)\odot\sigma\!\bigl(V_{2}h_{m}^{(n)}\bigr)\bigr]
\Bigr)}
{\displaystyle
\sum_{j=1}^{K^{(n)}}\!
\exp\!\Bigl(
w^{\mathsf T}\!\bigl[\tanh\!\bigl(V_{1}h_{j}^{(n)}\bigr)\odot\sigma\!\bigl(V_{2}h_{j}^{(n)}\bigr)\bigr]
\Bigr)}
\end{equation}
where \({w}\), \(V_{1}\), and \(V_{2}\) are learnable parameters. To ensure comparability across models, we spatially normalize all attention scores to a common patch size \(I_{2}\) (see Supplementary Information for details), and then fuse these normalized scores via a weighted sum to produce the final composite attention score:
\begin{equation}
    A_{m} \;=\; \sum_{n=1}^{N} w_{n}\,A_m^{(n)},
\qquad \text{s.t.}\;\; \sum_{n=1}^{N} w_{n}=1 \;
\end{equation}
Here \(A_{m}\) encodes a consensus of macro- and micro-level assessments from the entire MIL ensemble, balancing sensitivities across resolutions while suppressing model-specific noise. Patches are then ranked in descending order of \(A_{m}\), and only the top $p_1$-percent patches are forwarded to Stage 2 QHVAE-based cluster sampling for further refinement.

\noindent\textbf{Sampling Based on Cluster of Features from QHVAE.}
In the previous section, an attention-driven sampling strategy identified $p_1$-percent of discriminative patches.
In this section, we deploy a cluster-based sampling method to further distill semantic information and compress the patch set, boosting both computational efficiency and model performance in WSI analysis. 
The feature maps are extracted by the QHVAE model detailed in the preceding chapter, which are flattened and clustered using K-means into $k$ distinct clusters. Each of the clusters capture similar visual and medical characteristics. Let the cluster-centre vector be $\mathbf{C} = [c_1, c_2, \dots, c_k]$. The size of each cluster $c_i$ is computed, and the minimum cluster size $m_{\text{min}}$ is determined:
\begin{equation}
m_{\text{min}} = \min(c_1, c_2,…, c_k)
\end{equation}                                                  
Since patches within the same cluster exhibit semantic homogeneity, we uniformly select $m_{\min}$ patches from each cluster, ensuring a balanced and non-redundant representation. This method selects the top $p_{2}$-percent patches, which are then passed to the L2G‑Net described in the following section.

\subsection{Local-to-Global Network Model}

Prior work has concentrated on feature extraction from low-resolution patches, exhibiting suboptimal performance on high-resolution data. Our Local-to-Global Network (L2G‑Net) model, which incorporates the QHVAE introduced earlier, effectively captures high-fidelity features from high-resolution images for downstream aggregation.

\noindent\textbf{Multi--View Input.}  
To enable the reconstruction of high‐resolution images $I_3$, 
we decompose each input into two complementary views \cite{MVANet}: a distant view that captures global context and a close-up view that preserves fine‐grained details.
In the distant view, each image is downsampled using bicubic interpolation, producing the global information \(\textit{I}^{(G)},\) which provides holistic scene context.  In the close‑up view, the same image is uniformly partitioned into \(M\) nonoverlapping patches
$\{\textit{I}^{(L)}_m\}_{m=1}^M$, with \(M\) set to 4.

\noindent\textbf{L2G‑Encoder.}
In the local branch, as demonstrated in the previous section, the QHVAE achieves excellent performance on $I_1$ image feature–extraction tasks and can capture deep semantic information. Consequently, we reuse the pre‑trained QHVAE to obtain the \emph{local feature representation} of the $m$‑th patch $\textit{I}^{(L)}_{m}$:
\begin{equation}
	\textit{z}^{(L)}_{m}
	= \mathcal{E}_{\mathrm{QHVAE}}\bigl(\textit{I}^{(L)}_{m}\bigr)
\end{equation}
where \(\mathcal{E}_{\mathrm{QHVAE}}\) returns the latent feature $\textit{z}^{(L)}_{m}$ with deep semantic information. Meanwhile, in the global branch, the global view is processed by a Swin Transformer\cite{swinTransformer}, which is adept at capturing long-range dependencies, leveraging its hierarchical shifted-window attention for efficient feature aggregation:
\begin{equation}
	\textit{z}^{(G)}
	= \mathrm{SwinTransformer}(\textit{I}^{(G)})
\end{equation}
Local and global features are aligned via $\mathcal{T}_{m}$ and fused into unified features $\textit{z}^{(F)}$ , thereby capturing both fine-grained local details and the transformer-induced global contextual dependencies:
\begin{equation}
	\textit{z}^{(F)}
	= \textit{z}^{(G)}
	+\sum_{m=1}^{M}\mathcal{T}_{m}\bigl(\textit{\\z}^{(L)}_{m}\bigr)
\end{equation}
\noindent\textbf{L2G‑Decoder.}  
Starting from \(\textit{z}^{(F)}\) and given the remarkable reconstruction quality achieved by QHVAE, the decoder is deliberately kept simple, consisting of just upsampling and convolutional stages. To reconstruct the high-resolution image \(\hat{\textit{I}}\), for block \(i\):
\begin{equation}
\textit{z}^{(i+1)} 
= 
\mathrm{Conv}\bigl(\mathrm{Upsample}(\textit{z}^{(i)})\bigr)
\end{equation}
where \(\mathrm{Upsample}\) is interpolation, and \(\mathrm{Conv}\) is
convolution.
\noindent\textbf{Loss Function.}  
To enhance reconstruction fidelity across multiple image granularities, we introduce a hierarchical loss function grounded in the Manhattan norm:
Let the encoder produce the set of ground-truth images \(\{{\textit{I}}_k\}_{k=1}^4\), and the decoder generate the reconstructions \(\{\hat{\textit{I}}_{k}\}_{k=1}^{4}\). We define the total hierarchical loss, grounded in the \(L_1\) norm, as
\begin{equation}
\mathcal{L}_{\text{total}}
        = \sum_{k=1}^{4}
          \bigl\lVert
              \hat{\textit{I}}_{k}
              - \textit{I}_{k}
          \bigr\rVert_{1}
\end{equation}

\begin{table*}[ht]
  \centering
  \small
  \setlength{\tabcolsep}{1.9mm}
  \begin{tabular}{c ccc ccc ccc}
    \toprule
    \multirow{2}{*}{Method}
    & \multicolumn{3}{c}{Camelyon-16}
    & \multicolumn{3}{c}{TCGA-NSCLC}
    & \multicolumn{3}{c}{TCGA-RCC} \\
    \cmidrule(lr){2-4} \cmidrule(lr){5-7} \cmidrule(lr){8-10}
    & A.\% & F.\% & U.\% & A.\% & F.\% & U.\% & A.\% & F.\% & U.\% \\
    \midrule

CLAM     & 87.3±4.6 & 86.0±5.8 & 91.8±4.9  & 85.2±2.4 & 85.0±2.6 & 92.9±1.4 & 89.0±1.3 & 79.8±2.2 & 98.3±0.3 \\
\addlinespace
TransMIL & 85.3±2.7 & 83.3±3.1 & 90.3±2.0 & 80.5±2.6 & 79.1±4.2 & 80.4±2.8 & 90.0±1.5 & 86.2±2.1 & 97.3±0.7 \\
\addlinespace
PMIL     & 87.2±3.9 & 87.0±4.0 & 90.4±4.4 & \textbf{85.5±1.2} & 85.3±1.3 & 93.0±0.3 & 90.5±1.6 & 87.5±2.5 & 98.3±0.3 \\
\addlinespace
DTFD-MIL & 87.5±3.3 & 86.9±3.8 & 92.2±1.9 & 76.5±7.5 & 77.0±6.9 & 91.1±7.4 & 87.3±1.8 & \textbf{88.0±1.7} & 97.9±0.5 \\
\addlinespace
ROAM     & 82.9±3.9 & 82.2±3.5 & 92.0±2.1 & 85.2±2.6 & 85.1±2.7 & 92.5±0.9 & 89.7±1.0 & 87.6±1.0 & 97.3±0.5 \\
\addlinespace
ICMIL    & 85.0±4.2 & 83.4±4.7 & 89.9±3.6 & 85.3±1.3 & 83.8±1.8 & 92.8±0.3 & 87.0±1.6 & 84.4±2.3 & 94.9±1.1 \\
\addlinespace
FOCUS    & 77.5±4.0 & 76.8±4.5 & 84.8±6.1 & 84.5±1.8 & 84.3±1.8 & 92.0±1.8 & 89.5±2.6 & 86.8±2.5 & 98.1±0.1 \\
\addlinespace

    \midrule 
    Res.+Meanpool & 68.0±2.0 & 67.6±1.9 & 72.6±5.1
                  & 78.0±1.7 & 77.7±1.8 & 88.9±1.8
                  & 83.1±2.1 & 78.6±2.0 & 98.1±0.3 \\
\addlinespace
    Ours+Meanpool & 88.0±2.6$\uparrow$ & 88.1±2.5$\uparrow$ & 94.1±2.6$\uparrow$
                  & 83.3±2.7$\uparrow$ & 83.2±2.8$\uparrow$ & 91.3±1.2$\uparrow$
                  & 88.8±3.7$\uparrow$ & 85.6±4.9$\uparrow$ & 97.0±1.8$\downarrow$ \\
\addlinespace
    
    \midrule 
    \midrule 
    Res.+ABMIL    & 84.8±4.1 & 84.7±4.1 & 89.8±4.0
                  & 83.2±1.3 & 83.1±1.3 & 92.0±0.3
                  & 87.6±1.0 & 79.0±0.1 & 98.4±0.1 \\
\addlinespace
    Ours+ABMIL    & 89.5±4.0$\uparrow$ & 89.7±3.9$\uparrow$ & 95.1±2.4$\uparrow$
                  & 84.3±1.3$\uparrow$ & 84.3±2.5$\uparrow$ & 92.4±1.6$\uparrow$
                  & 90.2±5.3$\uparrow$ & 86.5±7.2$\uparrow$ & 97.2±1.9$\downarrow$ \\
\addlinespace
    
    \midrule
    \midrule 
    Res.+DSMIL    & 61.5±8.2 & 56.7±13.6 & 67.3±8.1
                  & 76.1±2.2 & 75.8±2.4 & 85.0±0.6
                  & 85.2±2.0 & 74.7±3.4 & 97.8±0.4 \\
\addlinespace
    Ours+DSMIL    & 88.0±3.5$\uparrow$ & 88.0±3.4$\uparrow$  & 95.1±1.8$\uparrow$
                  & 77.2±2.9$\uparrow$ & 77.0±3.1$\uparrow$  & 86.2±1.6$\uparrow$
                  & 86.5±6.1$\uparrow$ & 80.4±11.7$\uparrow$ & 96.6±2.1$\downarrow$ \\
\addlinespace
    
    \midrule
    \midrule 
    Res.+RRTMIL   & 88.3±1.4 & 87.3±1.0 & 94.5±2.2
                  & 82.7±3.7 & 82.3±4.2 & 93.3±1.9
                  & 89.2±5.0 & 85.6±0.5 & 98.1±5.0 \\
\addlinespace
Ours+RRTMIL   & \textbf{91.5±3.2$\uparrow$} 
              & \textbf{91.0±3.7$\uparrow$} 
              & \textbf{96.2±2.1$\uparrow$}
              & 85.4±1.4$\uparrow$ 
              & \textbf{85.3±1.5$\uparrow$} 
              & \textbf{93.5±0.9$\uparrow$}
              & \textbf{90.5±3.3$\uparrow$} 
              & 86.9±5.5$\uparrow$ 
              & \textbf{98.4±0.7$\uparrow$} \\

    \bottomrule

  \end{tabular}
    \caption{\normalfont Comparison with state-of-the-art methods (mean$\pm$std), where 'A.' denotes ACC, 'F.' denotes F1-score, and 'U.' denotes AUC. Our model is trained on only 4.5\% of the total training data, while all baseline models are trained on the entire dataset.}

  \label{tab:compared1}
\end{table*}

\section{Experiment}

\subsection{Experimental Setup}%
\noindent\textbf{Dataset.}
To comprehensively evaluate the performance of CDSR and compare it with state-of-the-art algorithms, we conduct experiments on three public datasets: Camelyon-16 \cite{Camelyon16}, TCGA-NSCLC, and TCGA-RCC. Camelyon-16 contains 270 training and 129 test whole-slide images. TCGA-NSCLC comprises 1,054 WSIs of lung squamous cell carcinoma (LUSC) and adenocarcinoma (LUAD), while TCGA-RCC includes 937 WSIs of clear cell, papillary, and chromophobe renal cell carcinoma. Camelyon-16 and TCGA-NSCLC serve as binary classification tasks, whereas TCGA-RCC is a three-class classification task. All datasets provide slide-level labels.

\noindent\textbf{Evaluation Metrics.}
We assess slide-level classification performance using three metrics: accuracy (ACC), F1-score, and area under the ROC curve (AUC). Additionally, we measure reconstruction accuracy using Peak Signal-to-Noise Ratio (PSNR) and Structural Similarity Index (SSIM).

\noindent\textbf{Compared Methods.}
We evaluate CDSR on the above datasets against state-of-the-art WSI classification methods, including ABMIL \cite{AB-MIL}, TransMIL \cite{TransMIL}, DTFD \cite{DTFD}, ROAM \cite{ROAM}, CLAM \cite{CLAM}, PMIL \cite{PMIL}, RRTMIL \cite{RRTMIL}, ICMIL \cite{ICMIL}, FOCUS \cite{FOCUS}, Mean-pooling, and DSMIL \cite{DSMIL}. All comparison models use a ResNet backbone as the feature extractor and are trained on approximately 1,003,152 patches. ICMIL and DSMIL employ self-supervised strategies, while the others use ImageNet-pretrained backbones.

\noindent\textbf{Implementation Details.}
In the CDSR framework, \textit{I\textsubscript{1}}, \textit{I\textsubscript{2}}, and \textit{I\textsubscript{3}} denote images of patch sizes 256, 1024, and 2048, respectively. Based on the patch sizes and the number of patches used, only 4.5\% of the total training data on average is utilized for L2G-Net training.  In the attention-based sampling stage, we select the top $p_1$-percent patches, with $p_1 = 5\%$ on Camelyon-16 and $p_1 = 8\%$ on TCGA-NSCLC and TCGA-RCC, considering the larger tumor regions in the TCGA datasets. During the clustering stage, we set the number of $k$-means clusters to 8. The TCGA-RCC dataset is also evaluated under the same experimental settings, with detailed results provided in the Supplementary material.

\subsection{Model Performance Analysis}

The performance of CDSR is evaluated from multiple perspectives. It achieves state-of-the-art results and demonstrates strong compatibility with various MIL models. Compared with other feature extractors, it maintains competitive performance while using substantially fewer training patches. Moreover, unlike self-supervised frameworks such as SimCLR \cite{simclr} and ICMIL, CDSR remains robust under limited data conditions. In addition, it offers significantly improved training efficiency over methods like DINO \cite{dino} and SimCLR. These results confirm the effectiveness and efficiency of CDSR.

\noindent\textbf{Comparison with Other Methods.}
Tab.~\ref{tab:compared1} presents the performance comparison of CDSR, when used as a feature extractor, against other baseline methods. Notably, CDSR can be flexibly integrated with various MIL aggregators. When combined with MeanPool and DSMIL, it achieves an average improvement of 24.6\% across all evaluation metrics on Camelyon-16, indicating the effectiveness of the extracted features, especially with relatively simple aggregation strategies. Additionally, CDSR attains state-of-the-art performance, delivering gains of 6.3\% in ACC, 5.5\% in AUC, and 7.7\% in F1-score, highlighting its predictive power and practical value.

Although CDSR does not fully surpass the SOTA in terms of ACC on TCGA-NSCLC and F1-score on TCGA-RCC, its performance remains within 1\% margin. Given that the model is trained on only 4.5\% of the total training data, this marginal difference underscores the efficiency of CDSR.

\begin{table}[ht]
  \centering
  \small
  \begin{tabular}{c cc cc}
    \toprule
    \multirow{2}{*}{Encoder}
    & \multicolumn{2}{c}{ Camelyon-16}
    & \multicolumn{2}{c}{ TCGA-NSCLC}\\
    \cmidrule(lr){2-3} \cmidrule(lr){4-5} 
    &  A.\% &  U.\% 
    &  A.\% &  U.\%  \\
    \midrule
    Res.IN      & 84.8±4.1 & 89.8±4.0
                & 83.2±1.3 & 92.0±0.3\\
    \addlinespace
    Res.SSL     & 85.3±6.8 & 92.0±3.3
                & 83.5±2.7 & 91.4±1.1\\
    \addlinespace
    CONCH       & 96.5±0.0 & 98.5±0.0
                & 93.3±2.5 & 97.5±1.1\\
    \addlinespace
   \textbf{CDSR}& 89.5±4.0 & 95.1±2.4
                & 84.3±2.5 & 92.4±1.6\\
    \bottomrule
  \end{tabular}
  \caption{\normalfont Performance comparison of different encoders on the Camelyon-16 and TCGA-NSCLC datasets. 'Res.IN' refers to ResNet-50 pre-trained on ImageNet, and 'Res.SSL' indicates ResNet-50 trained with self-supervised learning.}
  \label{tab:compared2}
\end{table}

\noindent\textbf{Comparison with Advanced Encoders.}
The CDSR framework is compared with representative encoders, including ResNet-50 pretrained on ImageNet (Res.IN), ResNet pretrained via self-supervised learning (Res.SSL), and the language model-based encoder CONCH \cite{conch}. All results are obtained using ABMIL as the MIL aggregator. As shown in Tab.~\ref{tab:compared2}, CDSR outperforms both ResNet-IN and ResNet-SSL across all evaluation metrics, achieving 89\% ACC and 95\% AUC on Camelyon-16 and 84\% ACC and 92\% AUC on TCGA-NSCLC. 

While CONCH achieves higher scores, it relies on large-scale pretraining over curated datasets and uses a considerably more complex architecture. In contrast, CDSR is lightweight and is typically trained on only a minimal subset of the total training data, highlighting its efficiency for scenarios with limited computational resources.

\begin{figure}[t]
\centering
\includegraphics[width=1\columnwidth]{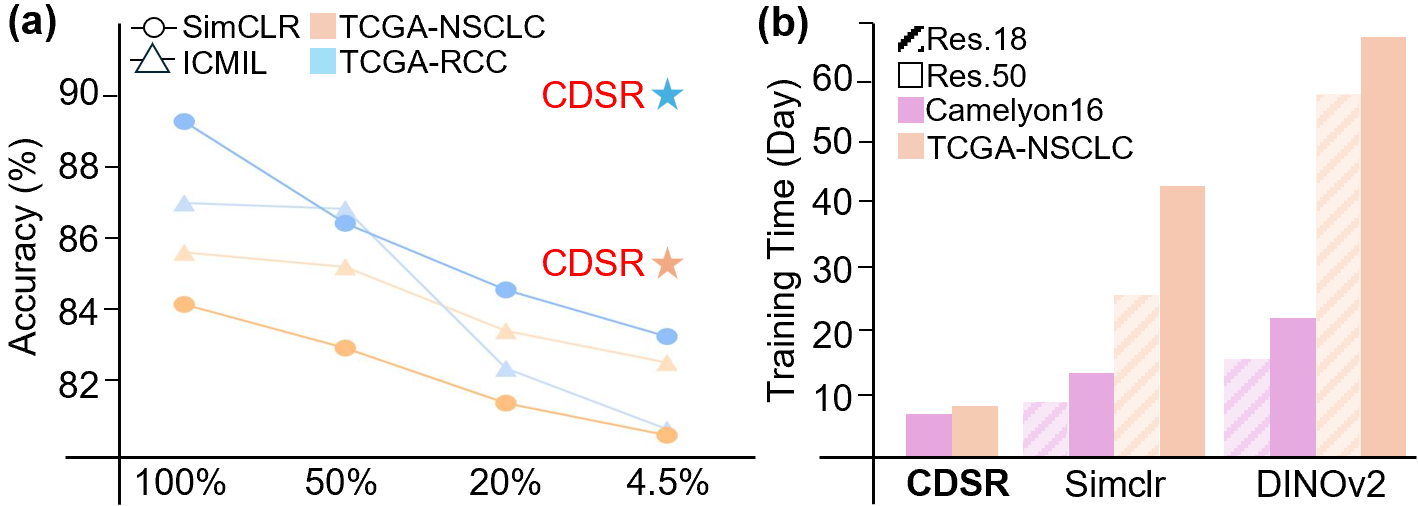}
\caption{Comparison of training data and training time.\\
(a) Accuracy of SimCLR and ICMIL under different training data proportions on TCGA-NSCLC and TCGA-RCC. 
(b) Training time comparison of DINO, SimCLR and CDSR on Camelyon-16 and TCGA-NSCLC.
}
\label{fig:figure3}
\end{figure}

\noindent\textbf{Comparison of Training Data.}
As illustrated in Fig.~\ref{fig:figure3}(a), the accuracy of CDSR is compared with two self-supervised frameworks, SimCLR and ICMIL, under varying training data scales. As the amount of training data decreases, both SimCLR and ICMIL exhibit performance degradation. In contrast, the CDSR framework achieves superior performance even with limited training data, demonstrating its strong data efficiency. This experiment is conducted on more heterogeneous TCGA-NSCLC and TCGA-RCC datasets.

Moreover, as shown in Fig.~\ref{fig:figure1}(c), CDSR outperforms ImageNet-pretrained baselines, including DSMIL, MeanPool, ABMIL, and RRTMIL. Notably, the accuracy of DSMIL improves with reduced data, likely due to its relatively simple architecture, which makes it more susceptible to redundancy in the training set and prone to overfitting.

\noindent\textbf{Comparison of Training Time.}
Fig.~\ref{fig:figure3}(b) compares the training efficiency of CDSR with two representative self-supervised frameworks: DINO and SimCLR. This experiment is processed on one NVIDIA A100 GPU. Training DINO and SimCLR for 100 epochs typically takes 2 to 10 weeks. In contrast, CDSR completes training in only 6.3 days on Camelyon-16 and 7.8 days on TCGA-NSCLC. These results underscore the superior efficiency of the CDSR framework.

\subsection{Effectiveness Analysis}
This section evaluates the effectiveness of the proposed CDSR framework from two aspects. The first experiment assesses the reconstruction quality of QHVAE and L2G-Net. The second experiment visualizes the spatial distribution of top $p_1$-percent patches in the attention-based sampling stage, illustrating the results under different values of $p_1$.

\begin{table}[ht]
  \centering
  \small
  \setlength{\tabcolsep}{1mm} 
  \begin{tabular}{c cc cc}
    \toprule
    \multirow{2}{*}{Encoder}
    & \multicolumn{2}{c}{ Camelyon-16}
    & \multicolumn{2}{c}{ TCGA-NSCLC}\\
    \cmidrule(lr){2-3} \cmidrule(lr){4-5} 
    
    &  PSNR &  LPIPS 
    &  PSNR &  LPIPS  \\
    \midrule
    QHVAE       & 45.60±2.18 & 0.003±0.00
                & 44.73±0.89 & 0.002±0.00\\
    \addlinespace
    L2G-Net     & 33.90±3.50 & 0.200±0.08
                & 29.33±2.17 & 0.198±0.05\\
    \bottomrule
  \end{tabular}
  \caption{\normalfont Reconstruction performance of QHVAE and L2G-Net on the Camelyon-16 and TCGA-NSCLC datasets. }
  \label{tab:table3}
\end{table}

\noindent\textbf{Reconstruction Quality Analysis.}
To evaluate reconstruction quality, QHVAE and L2G-Net are respectively tested on \(1024 \times 1024\) and \(2048 \times 2048\) patches. As shown in Tab.~\ref{tab:table3}, QHVAE achieves higher scores due to the smaller input size. Despite the increased difficulty from reconstructing larger patches, L2G-Net still achieves a competitive average PSNR of 29dB. These results demonstrate that both models are capable of producing high-quality reconstructions, with QHVAE achieving near-lossless quality, while L2G-Net preserves detailed structure even at larger scales. Reconstruction visuals are provided in the supplementary material.

\begin{figure}[t]
\centering
\includegraphics[width=1\columnwidth]{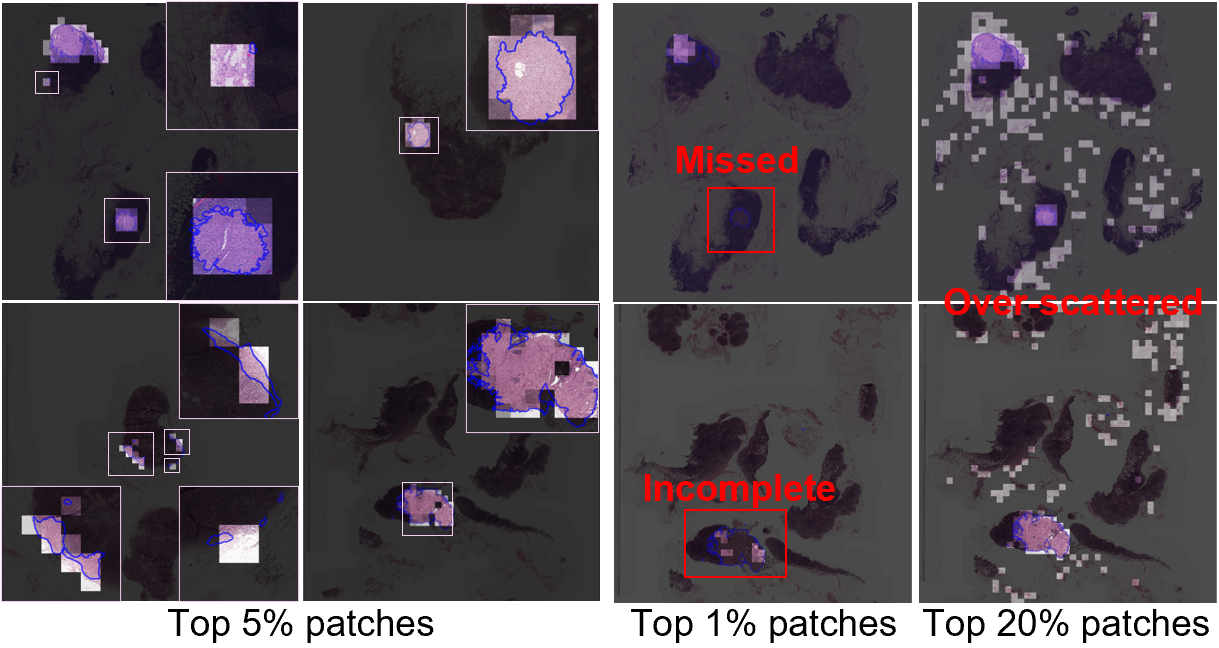} 
\caption{Visualization of top-$5\%$, top-$1\%$, and top-$20\%$ patch selections from WSI slides, ranked by attention scores. Ground-truth tumor annotations are overlaid for reference.}
    \label{fig:figure4}
\end{figure}

\noindent\textbf{Effect of Top-N Patches.}
To evaluate the effectiveness of top $p_1$-percent patches in the attention-based sampling stage, we visualize WSI slides on the Camelyon-16 dataset overlaid with the top-1\%, top-5\% (our selected $p_1$), and top-20\% patches, respectively. Note that this visualization is only conducted on Camelyon-16, as the TCGA-NSCLC and TCGA-RCC datasets do not provide pixel-level tumor annotations. As shown in Fig.~\ref{fig:figure4}, the blue contours represent the ground-truth tumor annotations. The top-1\% patches often provide incomplete coverage and may miss portions of the tumor region. The top-5\% patches exhibit comprehensive and focused tumor coverage, capturing most relevant regions with limited inclusion of irrelevant tissue. The top-20\% patches tend to be over-scattered and include substantial normal regions. These results indicate that the top-5\% patches offer an effective balance between sufficient tumor coverage and minimized inclusion of normal tissue.

\section{Conclusion}
In this study, we propose the Cascaded Dual-Scale Reconstruction (CDSR) framework to tackle two major challenges in WSI analysis. CDSR narrows the domain gap between natural and medical images that limits conventional MIL pipelines. Furthermore, it cuts information loss, redundant features, and computational overhead by curbing excessive patch splitting during self-supervised learning.
CDSR employs a two-stage selective sampling strategy to select only 9 informative patches per WSI on average. A Local-to-Global Network then delivers high-resolution reconstruction for robust representations.
Extensive experiments demonstrate that CDSR achieves improvements of 6.3\% in accuracy and 5.5\% in area under ROC curve with only 4.5\% of the total training data on average. These findings demonstrate that a handful of carefully chosen high-resolution patches can replace millions of low-resolution tiles without sacrificing diagnostic accuracy. In future work, we will extend CDSR to even larger patches or whole-tissue regions for true end-to-end learning, incorporate pathology-specific self-supervised objectives, and conduct prospective clinical validation.

\bibliography{aaai2026}

\end{document}